\documentclass[letterpaper, 10 pt, conference]{ieeeconf}  

\IEEEoverridecommandlockouts                              
\usepackage{graphics} 
\usepackage{amsmath}  
\usepackage{amssymb}  
\usepackage{subfig}
\usepackage{graphicx,epsfig,url}
\usepackage{multirow}
\usepackage[normalem]{ulem}
\usepackage{placeins}
\usepackage{floatrow}
\usepackage{placeins}
\usepackage{hyperref}
\usepackage{ifthen}
\newboolean{ABR}
\setboolean{ABR}{false}

\usepackage{xspace}

\newcommand{\changed}[1]{\textcolor{black}{#1}}

\title{\LARGE \bf 
Coarse-to-fine Surgical Instrument Detection for Cataract Surgery Monitoring
}

\author{Hassan Al Hajj, Gwenol\'e~Quellec, Mathieu~Lamard, Guy~Cazuguel, B\'eatrice~Cochener 
\thanks{H. Al Hajj, G. Quellec, M. Lamard, G. Cazuguel and B. Cochener are with Inserm, UMR 1101, Brest F-29200, France
{\tt\small hassan.alhajj@inserm.fr}}%
\thanks{M. Lamard and B. Cochener are with Univ Bretagne Occidentale, Brest, F-29200 France}%
\thanks{G. Cazuguel is with INSTITUT TELECOM; TELECOM Bretagne; UEB; Dpt ITI, Brest, F-29200 France}%
\thanks{B. Cochener is with Service d'Ophtalmologie, CHRU Brest, Brest, F-29200 France}
}

\begin{document}

\maketitle
\thispagestyle{empty}
\pagestyle{empty}

\begin{abstract}
The amount of surgical data, recorded during video-monitored \changed{surgeries}, has extremely increased. This paper aims at improving existing solutions for the automated analysis of cataract surgeries in real time. Through the analysis of a video recording the operating table, it is possible to know which instruments exit or enter the operating table\changed{, and therefore} which ones are likely being used by the surgeon. Combining these \changed{observations} with \changed{observations from} the microscope video should enhance the overall performance of the systems. To this end, the proposed solution is divided into two main parts: one to detect the instruments at the beginning of the surgery and one \changed{to update the list of instruments} every time a change is detected in the scene. In the first part, the goal is to separate the instruments from the background and from irrelevant objects. For the second, we are interested in detecting the instruments that appear and disappear \changed{whenever the surgeon interacts with the table}. Experiments on a dataset of 36 cataract surgeries validate the good performance of the proposed solution.
\end{abstract}

\section{Introduction}

\begin{figure}[!h]
  \begin{center}
    \begin{tabular}{cc}
      \subfloat[Operating table]{
        \includegraphics[width=.44\textwidth]{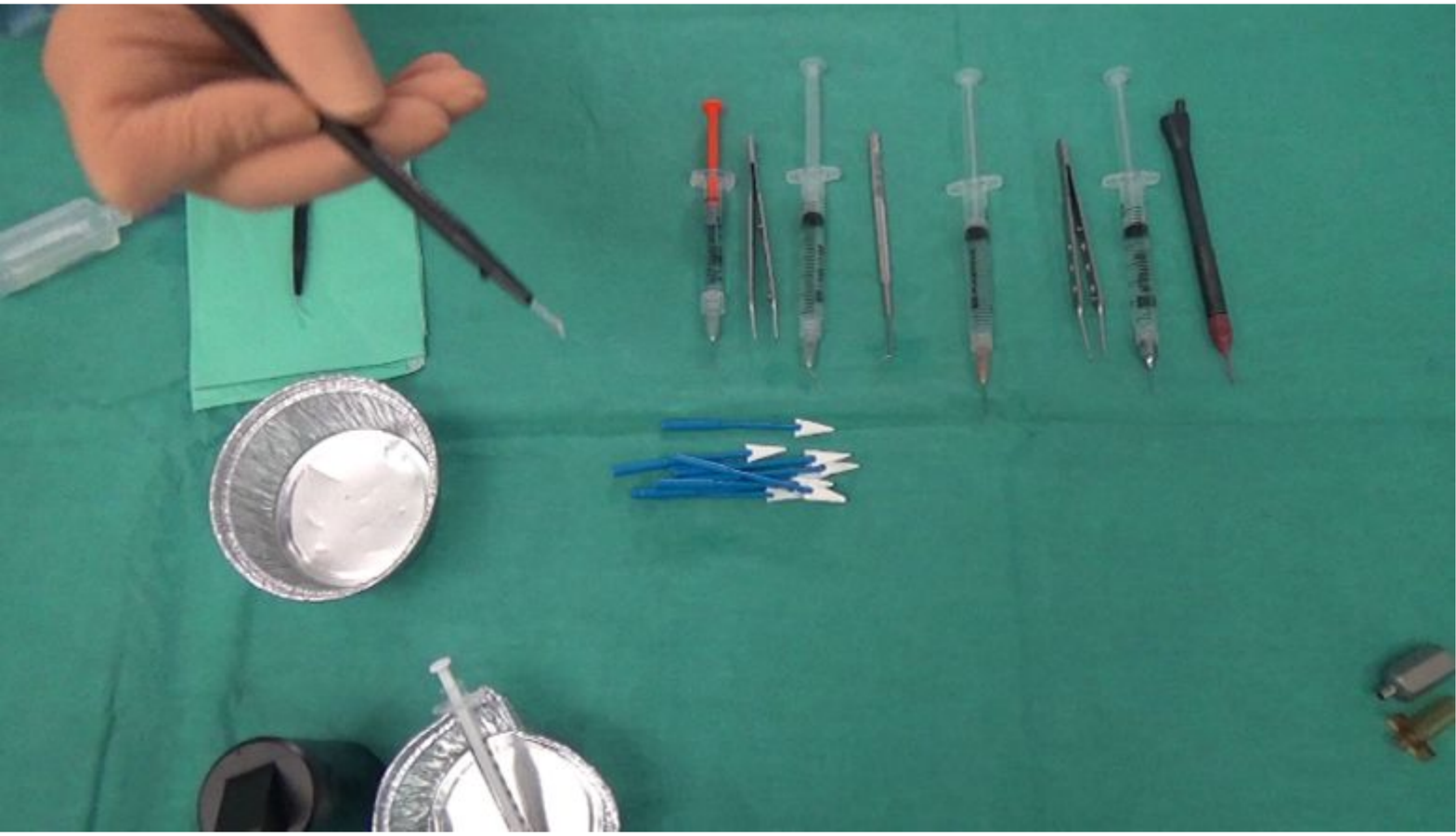}
       } &
      \subfloat[Microscope field of view]{
        \includegraphics[width=.44\textwidth]{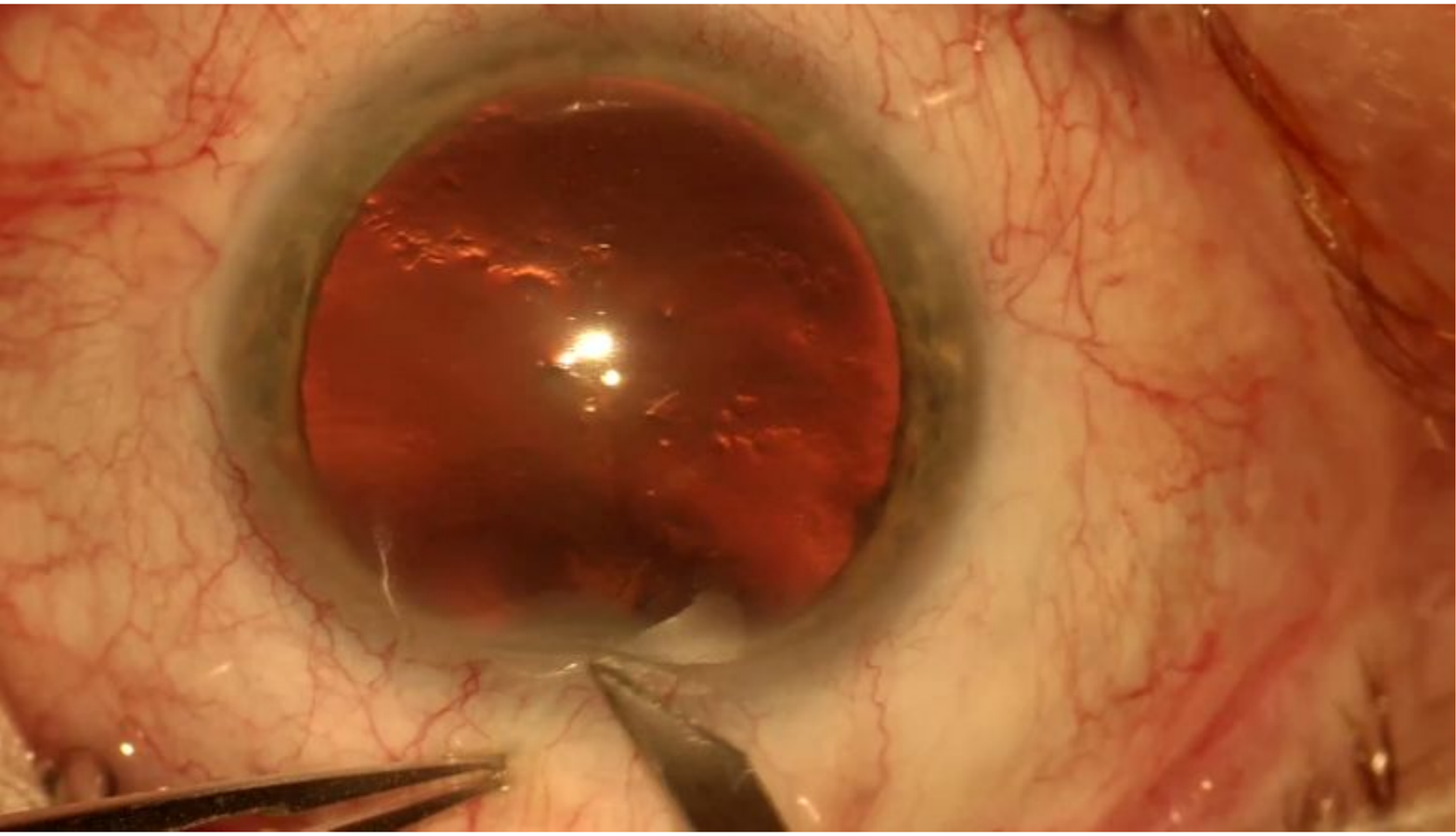}
      }
    \end{tabular}
  \end{center}
  \caption{Operating table image captured at $t$. Microscope image captured at $t$ \changed{+ a few seconds}, showing part of the knife that has been taken out from the table.}
  \label{fig:operatingTableRelationMicroscope}
\end{figure}

With the emergence of many medical imaging devices and technologies in the operating rooms (MRI, ultrasound imaging, surgical microscope, etc .), the automated analysis of data recorded during video monitored surgeries has developed during the last decade. \changed{Methods} emerged in this research field could help the surgeons in different manners: report generation \cite{lalys_framework_2012, stanek_automatic_2012}, surgical skill evaluation or construction of \changed{educational} videos \cite{andre_learning_2012}. Also, real-time video monitoring \changed{would allow automatically communicating useful} information \changed{to} the surgeon during the surgery.\\
\changed{For instance,} studies have been initiated to setup warning/recommendation generation \changed{systems} for video monitored surgery. \changed{This includes fast and robust methods to recognize surgical tasks, steps or gestures in real time \cite{charriere_automated_2014, quellec_real-time_2014, quellec_real-time_2015}.} \changed{With such methods, it} would be possible to distinguish a normal conduct of surgery from an abnormal one. The results obtained are very encouraging, but highlighted \changed{one main challenge:} to improve the interpretation of the video, \changed{one should be able to detect} all surgical instruments. But, these instruments have a wide variety of shapes and are \changed{only} partially visible in the surgical scene. A lot of studies tackled the surgical instrument detection problem. The work carried out can be divided into two categories. The first category is the identification by radio frequency methods (RFID) \cite{surgical_instruments_automatic_identification}. The second category is based on image processing. Compared to the first category, the biggest advantage of the image processing methods is that they do not require any installation of additional components in the operating room \changed{that would alter} the surgical procedure.\\ 
To solve the partial occlusion problem, we propose the addition of a second video stream, filming the operating table (see Fig. \ref{fig:operatingTableRelationMicroscope}). By knowing which instruments exit or enter the operating table we know which tools are likely being used by the surgeon and which tools surely are not. In this context, a lot of methods were proposed for detecting, monitoring and recognizing the surgical instruments in different areas of medical surgery
\cite{minimally_invasive_instrument_detection, baldas_real-time_2010, retinal_microsurgery_instrument}. All these methods have focused on a small number of highly differentiated instruments \cite{surgical_instruments_classification_system_2014}. We work in a different context: we are dealing with many instruments, many of which resemble strongly. Although instruments are more easily detected on the surgical table than in the microscope video, analyzing the table video is challenging as well, due to the variety of actions that can be realized by the surgeons on the operating table (preparing implant, \changed{filling} in the syringes, etc.).\\
In this paper, we present two methods: one to segment the instruments at the beginning of the \changed{surgery} and one to detect the instruments that appear and disappear along the \changed{surgery}.

\section{Method Overview}
We have two objectives, i.e. two tasks to accomplish: \changed{1) describing the initial state, at the beginning of the surgery and 2) describing changes, whenever motion is detected on the table.} The reason we propose to describe the changes, rather than rerunning state description every time a change is detected, is that we assume change detection is more accurate \changed{in view of} the issues we may encounter in the table scene, e.g. occlusion problems. \\
A similar solution will be proposed for both tasks. Because there are many instruments, we do not want to manually define a model for each tool. Instead, we propose a general, strongly supervised solution. In that purpose, manual ground truth was provided for a subset of images from the video dataset. For the first task, instruments have been manually segmented. For the second task, changes (appearing/disappearing instruments) have been manually segmented. The solutions are based on analogy reasoning (k-NN): images are divided in patches. A feature vector is extracted for each patch: they describe the local visual content (for the first task) or the local change (for the second task). Using the manual segmentation associated with the nearest neighbors, an instrument probability (for the first task) or a change probability (for the second task) is computed.\\
Several solutions are proposed to speed up computations: images are downsampled by a factor of two, a fast approximation of k-NN is used \cite{approximate_nearest_neighbor_2009} and a coarse-to-fine search strategy is proposed.

\section{Static Instrument Detection} 

In this section, we handle the first task, called initial state description. \changed{It} is necessary to note that there are no motions by any means, no hands are hiding any part of the scene and no tasks are taking placed on the table. Then, it is about separating the instruments from the background.

\subsection{Challenges}
The tablecloth color is obviously uniform (\changed{green} color) and easily differentiable from the color range of the instruments. However, the background contains more objects than just the tablecloth: it contains all objects that are not surgically relevant, such as the piece of towel in Fig. \ref{fig:staticDetectionResults}\subref{fig:staticImageWithTowel}. In fact, the background contains all objects that the expert did not manually segment in the reference images, hence the relevance of the proposed strongly-supervised solution.

\subsection{Patch Description}
\label{patchDescription}
Simple visual features are proposed in this study. \changed{For each patch,} we extracted the \textit{mean} and the \textit{standard deviation} of the intensity values of the \textit{R}, \textit{G}, \textit{B}, \textit{H}, \textit{S} and \textit{V} \changed{channels}, in addition to the \textit{mean} and the \textit{standard deviation} of the result of \textit{Sobel} edge detection applied \changed{to the luminance channel}. It results in a vector descriptor of 14 elements.

\subsection{Cross-validation}
\changed{The system is trained and tested using leave-one-out cross-validation. While processing an image (the test image), all other images are used as references.} For each patch in the reference dataset, an instrument probability is computed: it is defined as the percentage of pixels inside the patch that were manually segmented by the expert.
\subsection{K-NN Regression}
Given a patch in the test set, the k nearest neighbors from the reference set \changed{are searched for}: the patch probability is defined as the average instrument probability among the nearest neighbors.
\subsection{Coarse-to-fine Strategy}
For faster computations, \changed{we propose to} start with large patches and subdivide them if \changed{and only if} instrument probability is greater than \changed{0\%} and so on until the desired patch size is reached.
\subsection{Parameter Estimation}
We introduced 4 parameters for this part. $K$ indicates the number of the nearest neighbors to be taken into consideration. $P_{min}$ is the smallest patch size in the list of \changed{patch sizes}. $\tau$ is the scale factor used to go from a scale level to another and last but not least $P$-levels is the number of the scale levels to be run. To find the optimal value for these parameters, which are discrete, a discrete version of the \textit{Particle Swarm Optimization} (PSO) algorithm was used here, called D-PSO \cite{datta_real-integer-discrete-coded_2011}.

\section{Dynamic Instrument Detection}
In this section, we are interested in detecting the instruments that appear and disappear along the way. In other words, detecting, at every moment, the instruments that are probably in the microscope scene. In this study, we compare the last image before a motion is detected in the table scene (the 'before' image) \changed{to} the first image after motion stops (the 'after' image).
\subsection{Challenges}
The surgeon does not simply put one instrument on the table and/or take another one. First, the surgeons usually moves several instruments around to search for the right instrument. Second, they use some of the instruments \changed{to accomplish} some tasks over the table, e.g. preparing implants. Therefore, many instruments are displaced without going out of the scene or used in the surgery. Then, the main challenge is to differentiate instruments that were simply moved around from instruments that have appeared on or disappeared from the table.

\subsection{Appearance and disappearance detection}
Without loss of generality, we only focus on appearance detection. To detect appearance in one patch from the 'after' image, the corresponding patch in the 'before' image is selected and these patches are compared. To detect disappearance, we simply swap the 'before' and 'after' images and run the analysis again.
\subsection{Compensating For Instrument Motions}
Considering the fact that the instruments are being displaced over the table, it is most likely possible that a patch \textit{P1} at position \textit{X} in the 'after' image will be found at position \textit{X + l} in the 'before' image, as a patch \textit{P2}, where \textit{l} is the displacement distance. Patch P2 is searched for inside a window centered on P1. P2 is defined as the patch whose feature vector V2 minimizes the Euclidean distance with V1, the feature vector extracted from P1.
\subsection{Change Description}
We extracted the same features used in the first task, detailed in section \ref{patchDescription}. In this part, the change is described by the difference between feature vectors (V2-V1). In case of instrument appearance, no match will be found in the 'before' image, so the difference will be large. If case of instrument motion, the difference will be close to zero.
\subsection{Analogy Reasoning}
For each patch in the reference \changed{'after'} images, we computed the change description, which implies looking for the most similar patch in the 'before' images. The cross-validation, k-NN regression and coarse-to-fine strategy are similar to the first task but one more parameter has been added to this part. $W$-size is the window size in which we look up for the best match in the 'before' image.
\section{Experiments}
\subsection{Cataract Surgery Dataset}
\subsubsection{Data Collection}
a dataset of 36 cataract surgery videos, recorded at Brest University Hospital between February and September 2015, were used in this experiment. These surgeries were carried out by two different surgeons. Each surgery is recorded \changed{as} two videos, one for the operating table and the other \changed{one} for the microscope field of view. Videos were acquired in full HD  pixel resolution and a frame rate of 50 FPS for the former and 30 FPS for the latter.
\subsubsection{Static Method Ground Truth}
to be able to detect the instruments \changed{statically}, we captured 36 frames: \changed{one frame at the beginning of each table video}. They were segmented manually by delineating the boundaries of all the instruments \changed{visible on} the table. Examples of images that were manually segmented are given in  Fig. \ref{fig:staticDetectionResults}\subref{fig:staticSegmentedImageWithoutTowel}\subref{fig:staticSegmentedImageWithTowel}. 
\subsubsection{Dynamic Method Ground Truth}
\changed{to detect the instruments dynamically}, 36 surgeon actions were selected randomly, one per video. An action is considered as an act of taking out an instrument from the table, putting it back or both at the same time. 2 images were captured for each action, one right before it, the other \changed{one} right after it. Those images were manually segmented by delineating the boundaries of the instruments that appeared and disappeared along the way. \changed{Instruments} that were simply displaced were not segmented. Examples of images that were manually segmented are given in  Fig. \ref{fig:dynamicDetectionResults}\subref{fig:objectDectectionManualSegmentation1}\subref{fig:objectDectectionManualSegmentation2}.

\subsection{Results \changed{and Discussion}}
\subsubsection{Static Method}
\begin{figure}
  \begin{center}
    \begin{tabular}{cc}
    \subfloat[Original image]{
        \includegraphics[width=0.44\textwidth]{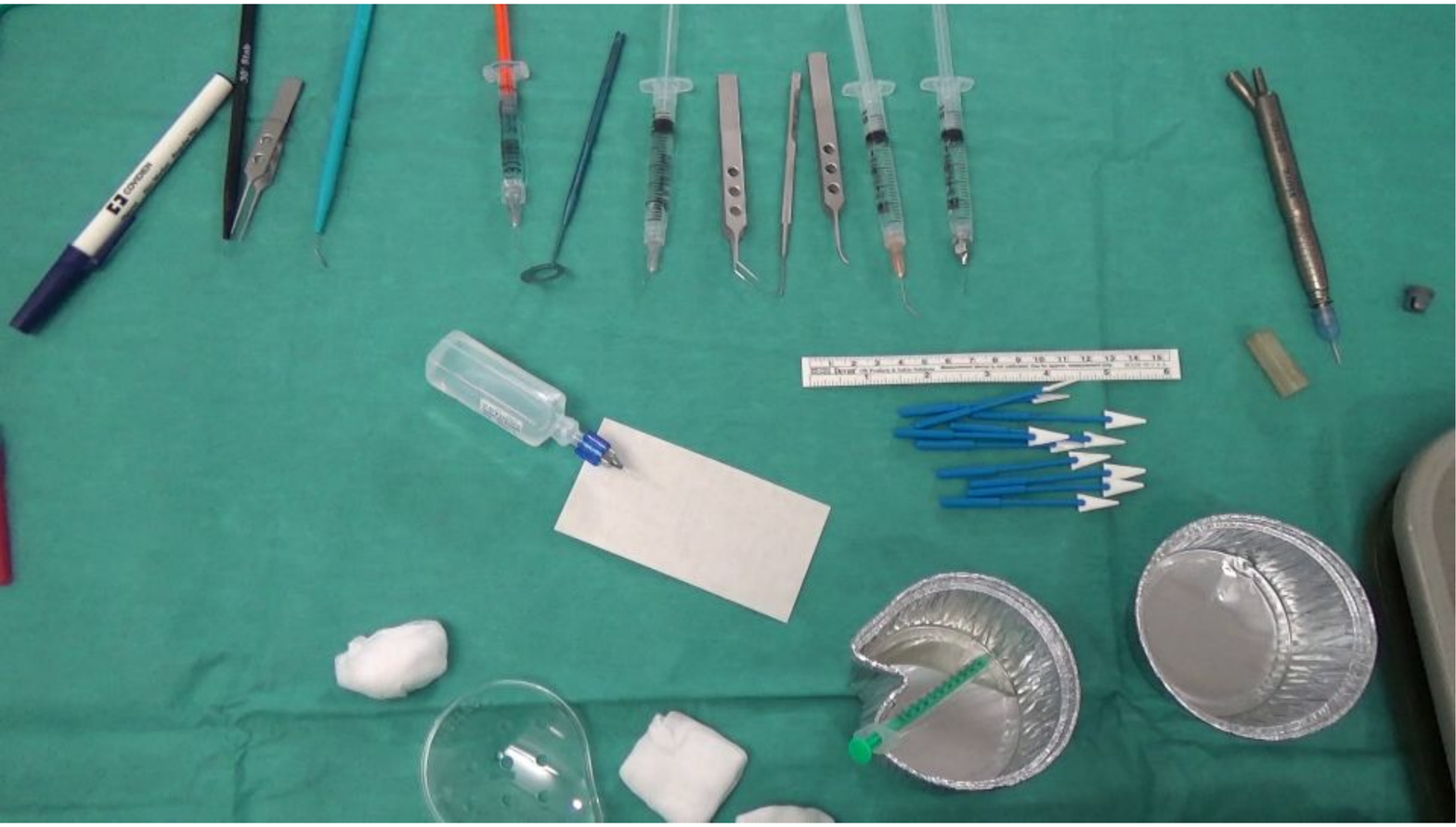}
      } &
      \subfloat[Original image]{
        \includegraphics[width=0.44\textwidth]{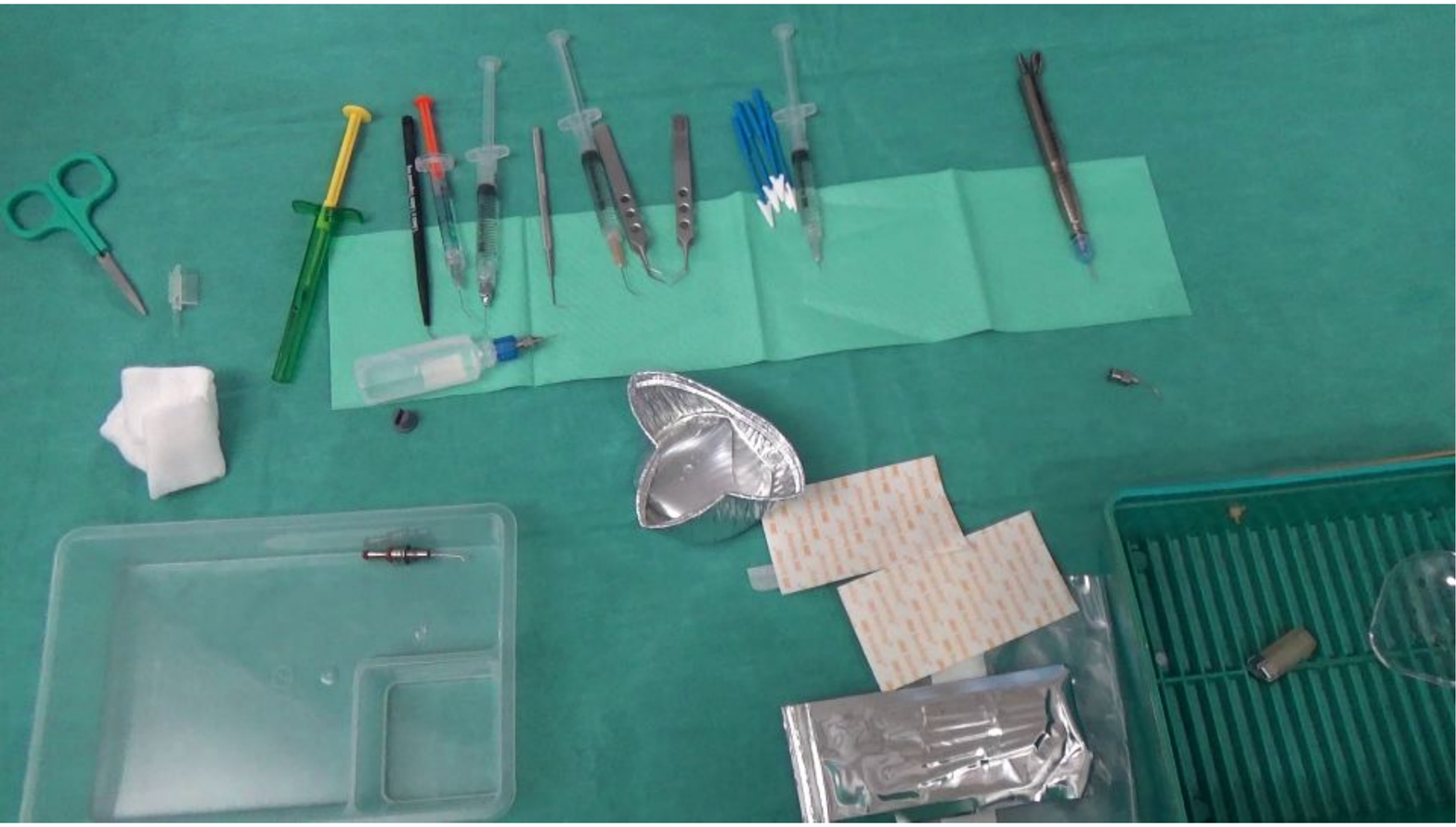}
        \label{fig:staticImageWithTowel}
      } \\
	  \subfloat[Image segmented manually]{
        \includegraphics[width=.44\textwidth]{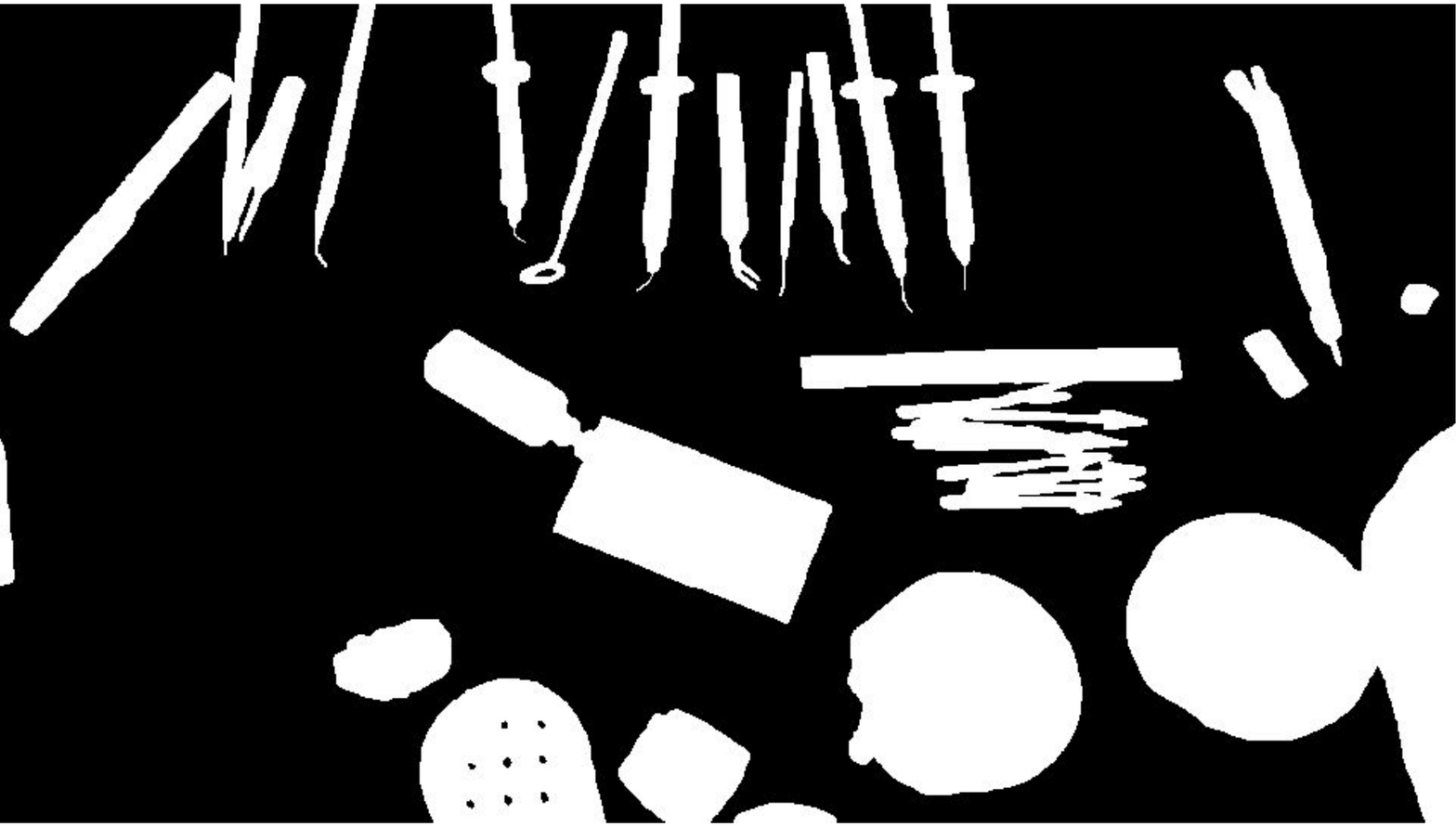}\label{fig:staticSegmentedImageWithoutTowel}
      } &
      \subfloat[Image segmented manually]{
        \includegraphics[width=.44\textwidth]{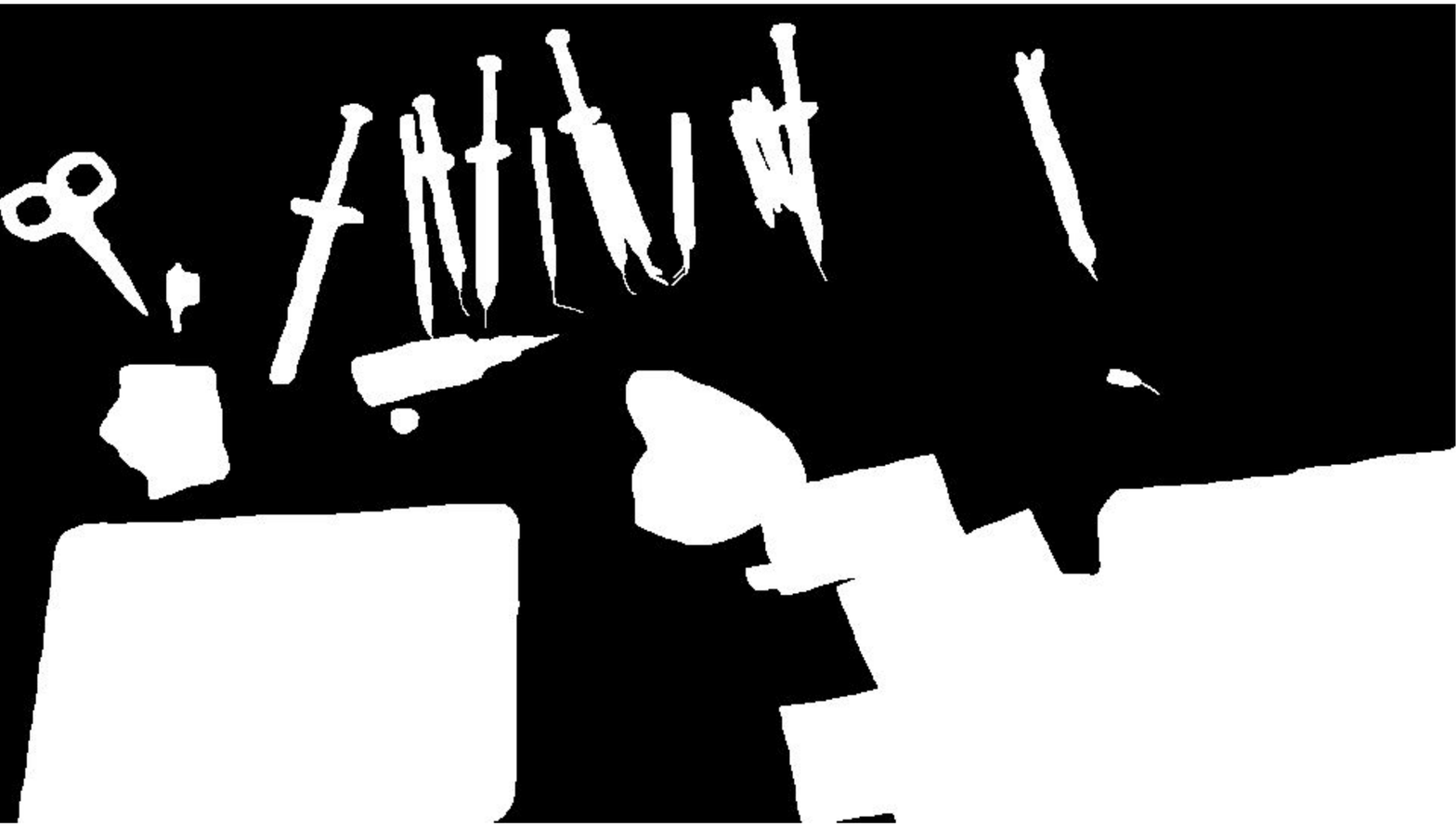}\label{fig:staticSegmentedImageWithTowel}
      } \\
	\subfloat[Result of  instrument detection]{
        \includegraphics[width=.44\textwidth]{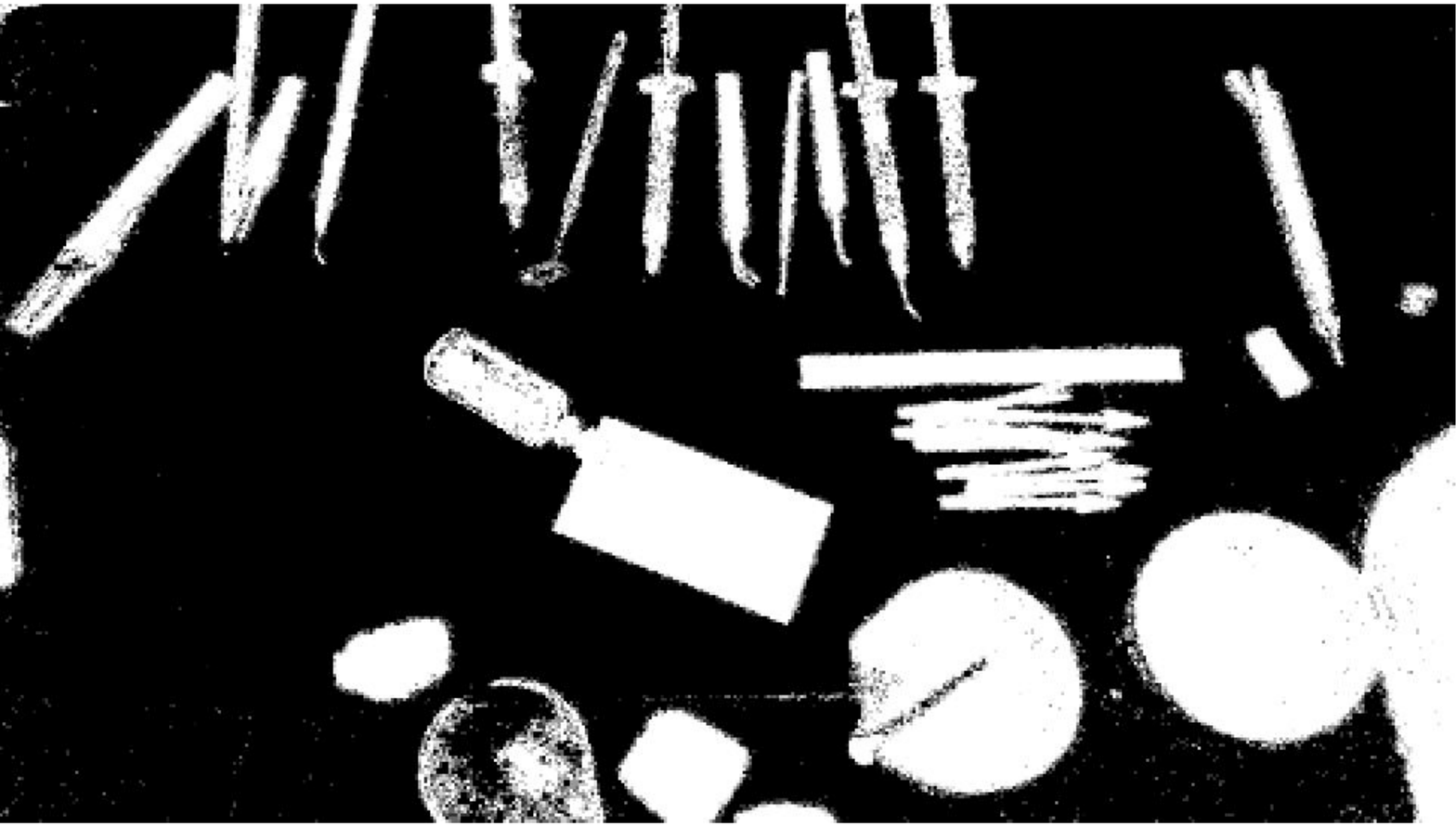}
      } & 
      \subfloat[Result of instrument detection]{
        \includegraphics[width=.44\textwidth]{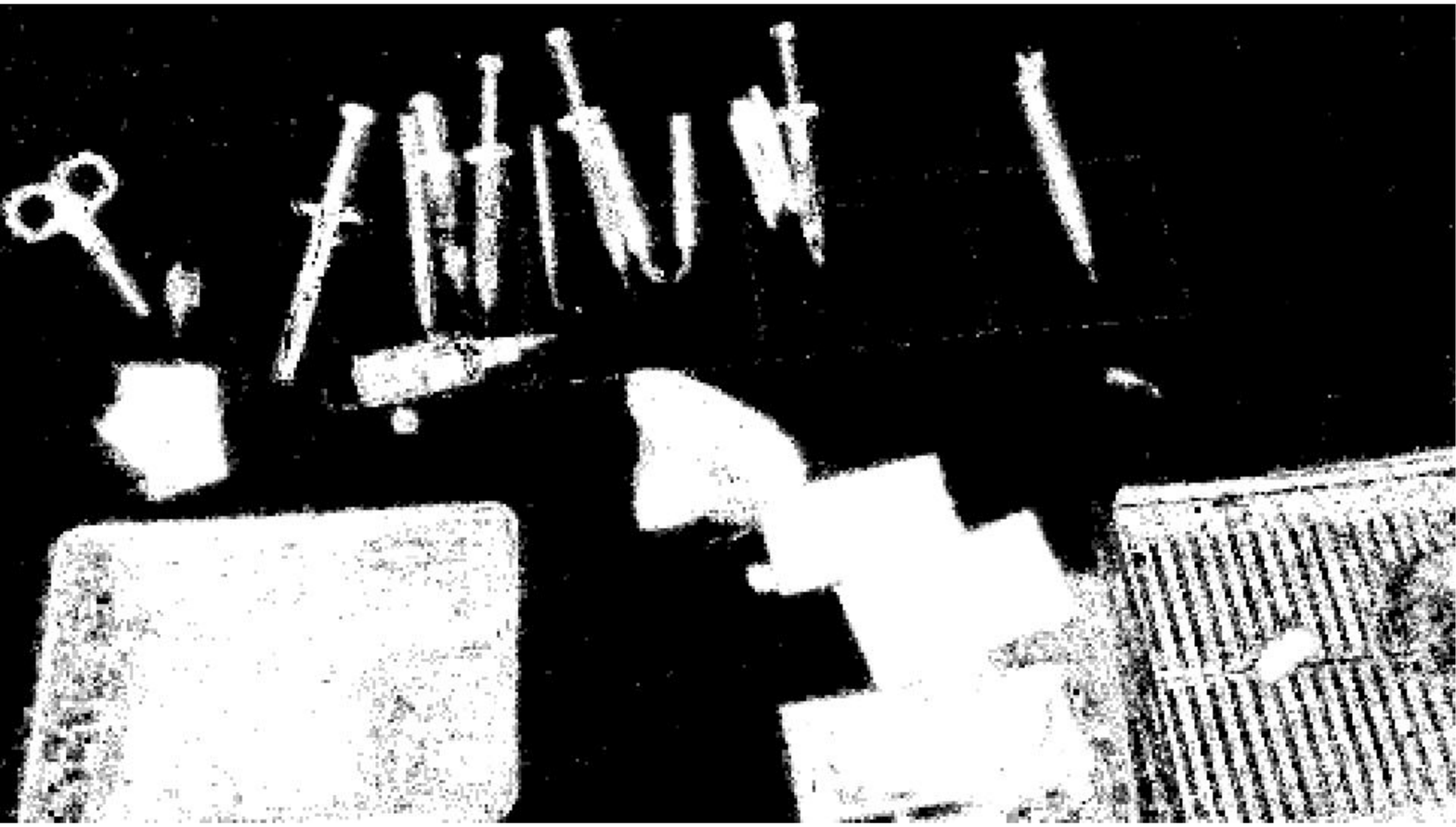}
      }	
    \end{tabular}
  \end{center}
  \caption{ Two examples of static instrument detection.}
  \label{fig:staticDetectionResults}
\end{figure}

algorithm parameters were optimized using D-PSO. The performance of the system is measured for each set of parameters in terms of Az, the area under the \textit{Receiver Operating Characteristic} (ROC) curve. The results are presented in Table \ref{tab:azStaticInstrumentDetection}. They show that we could strongly separate the instruments from the background as we can see in Fig. \ref{fig:staticDetectionResults}.
We can also see in Fig. \ref{fig:staticDetectionResults}\subref{fig:staticSegmentedImageWithTowel} that the towel, which was not segmented in reference images, was not segmented in the test image neither. Conversely, the large greenish and transparent containers, which the expert decided to segment in reference images, were segmented in the test image as well.


\begin{figure}
  \begin{center}
    \begin{tabular}{cc}
      \subfloat[Image before action]{
        \includegraphics[width=.44\textwidth]{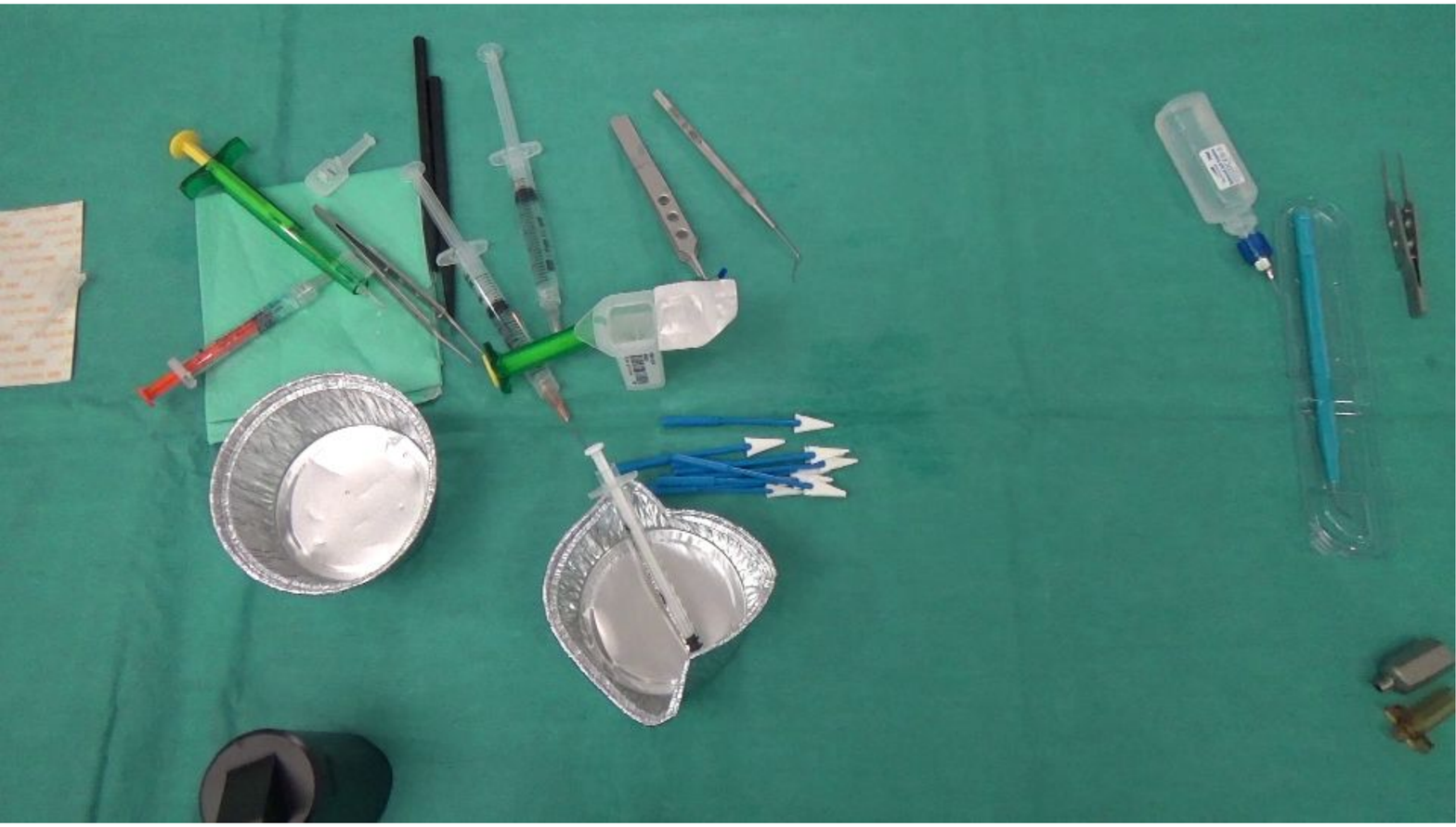}
      } &
      \subfloat[Image after action]{
        \includegraphics[width=.44\textwidth]{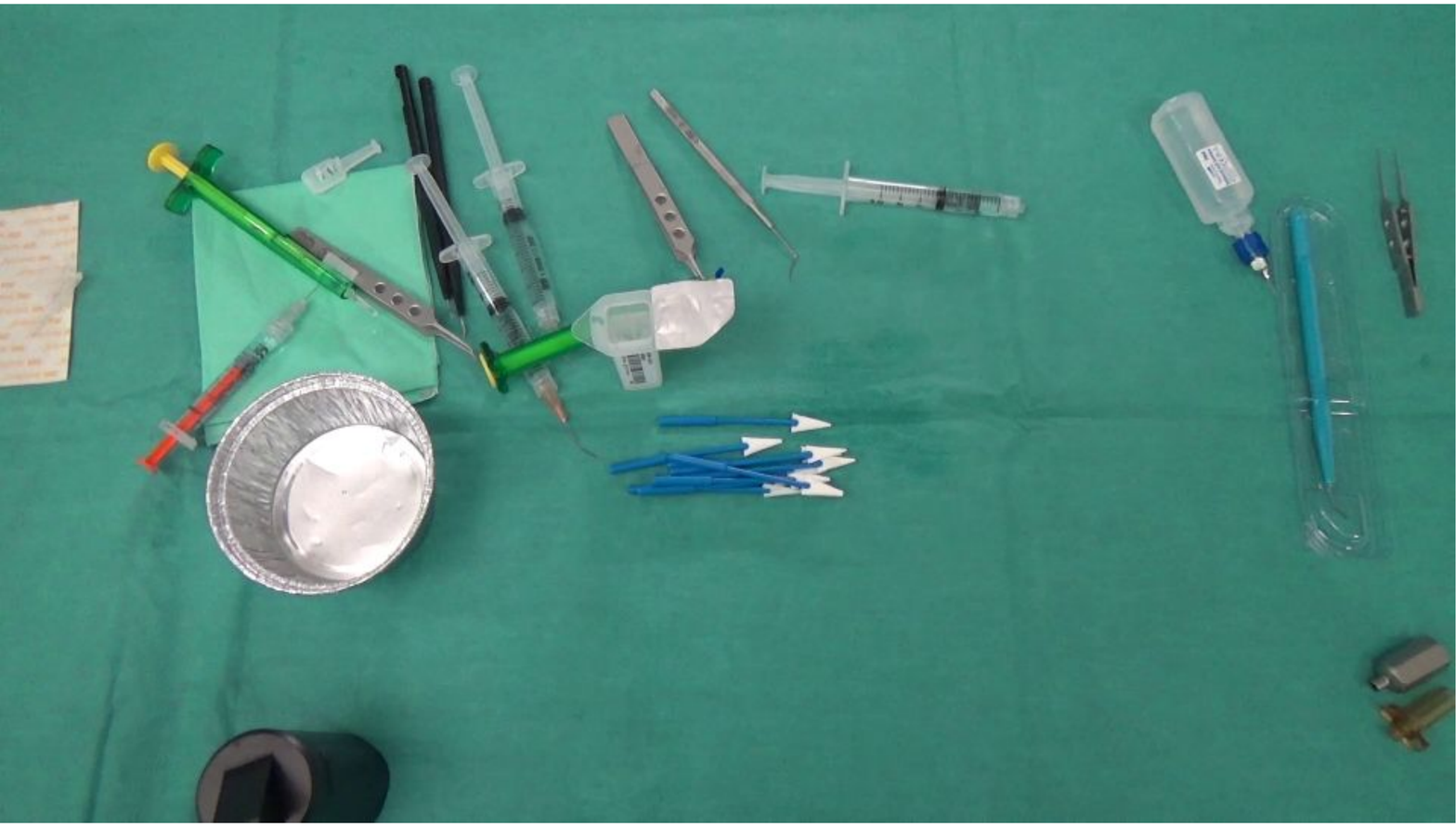}
      } \\
      \multicolumn{2}{c}{\subfloat[Difference between the 'after' and 'before' images]{\includegraphics[width=0.44\textwidth]{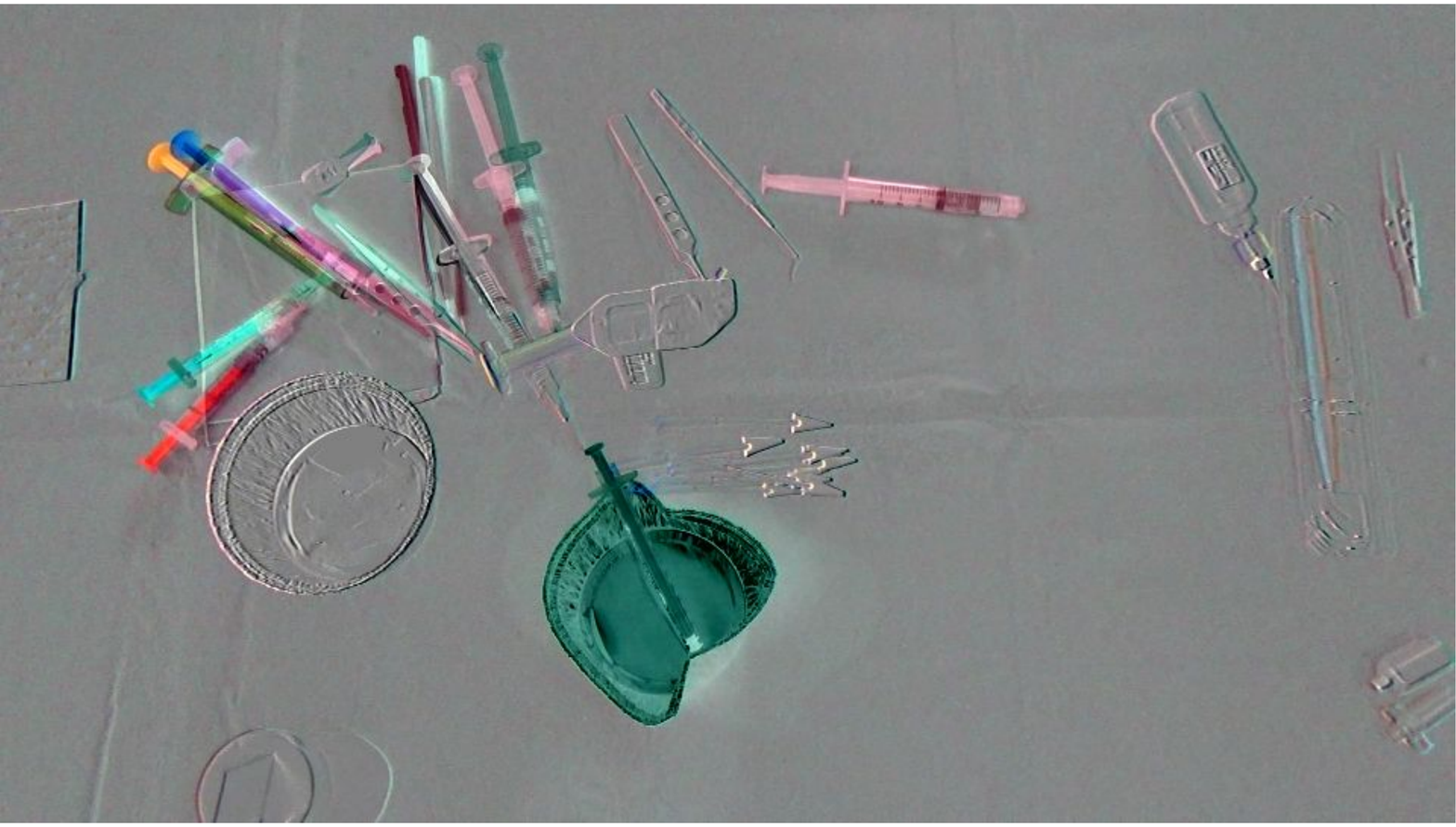}}}\\
      \subfloat[Manual segmentation of \changed{'before' and 'after' images}]{
        \includegraphics[width=.44\textwidth]{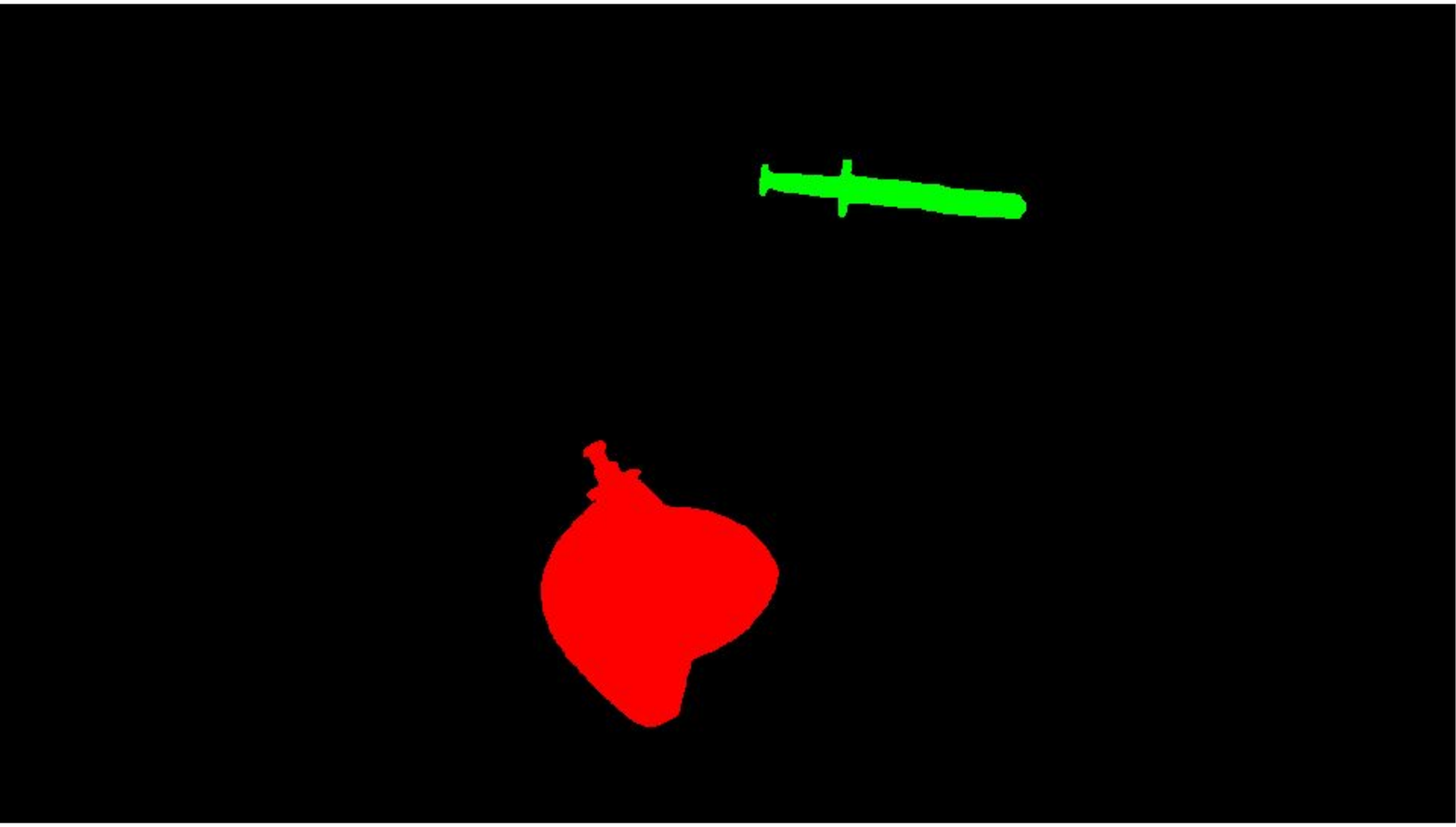}\label{fig:objectDectectionManualSegmentation1}
      } &
      \subfloat[Result of instrument detection]{
        \includegraphics[width=.44\textwidth]{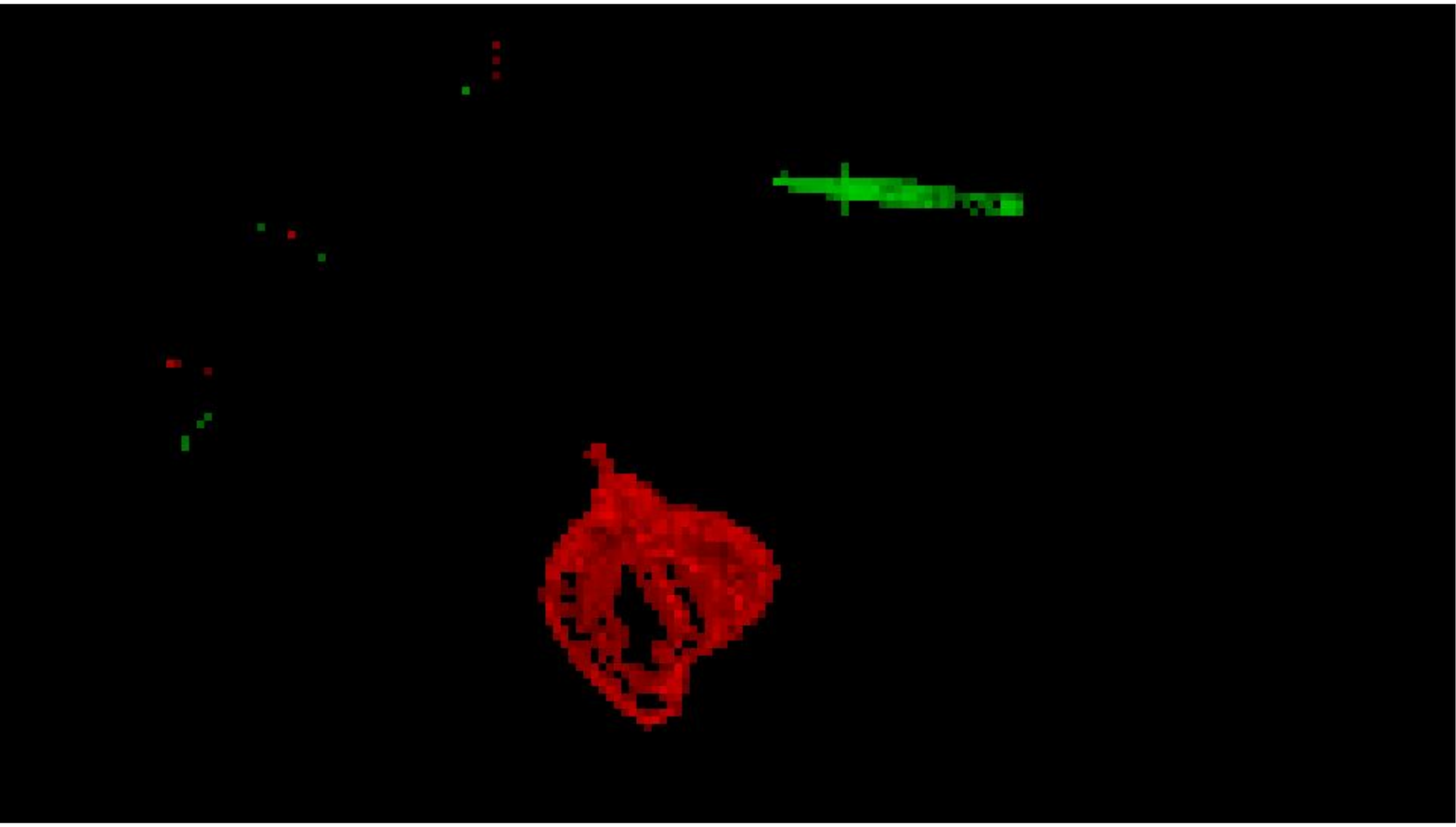}
      } \\
      \subfloat[Image before action]{
        \includegraphics[width=.44\textwidth]{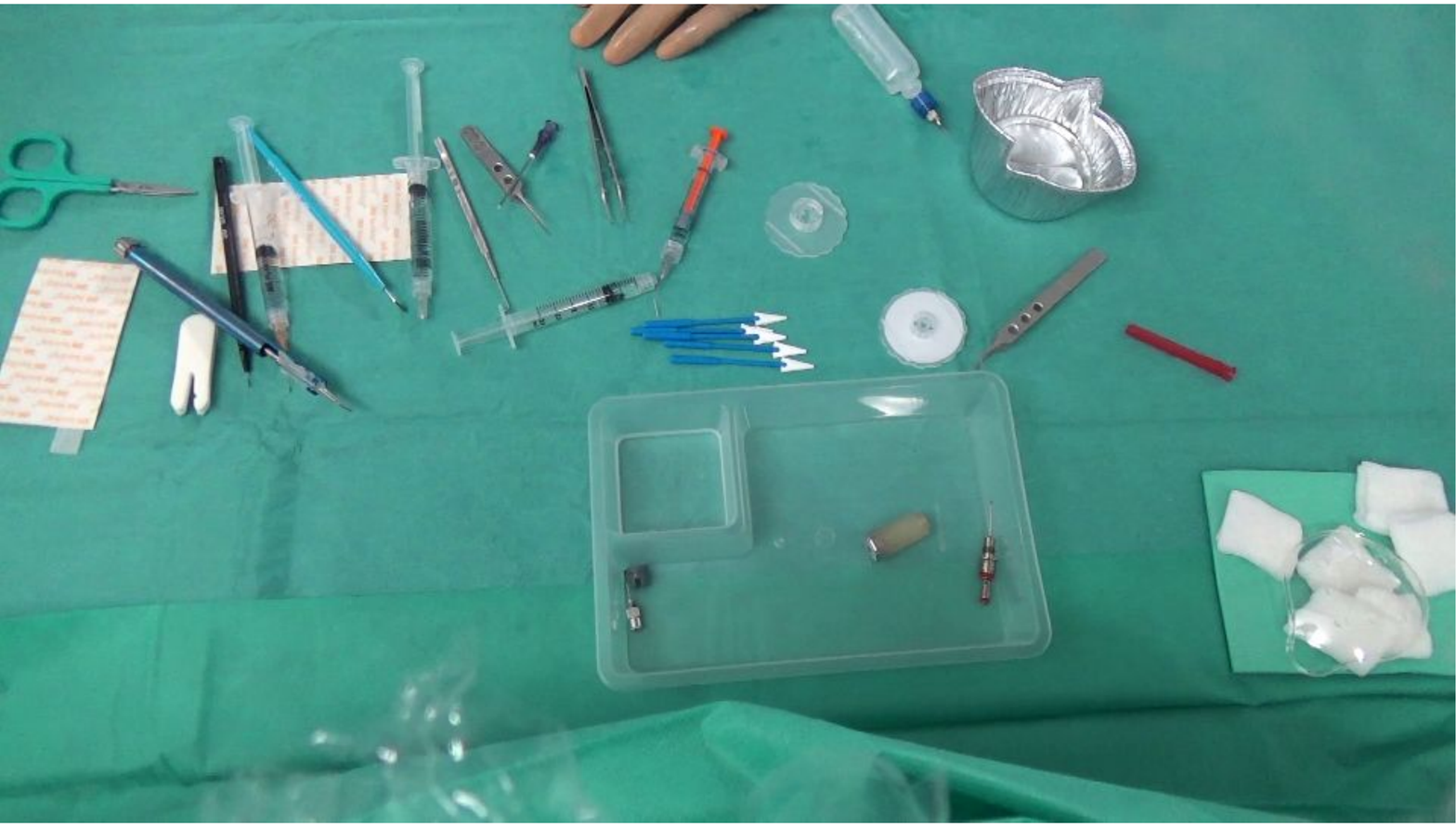}
      } &
      \subfloat[Image after action]{
        \includegraphics[width=.44\textwidth]{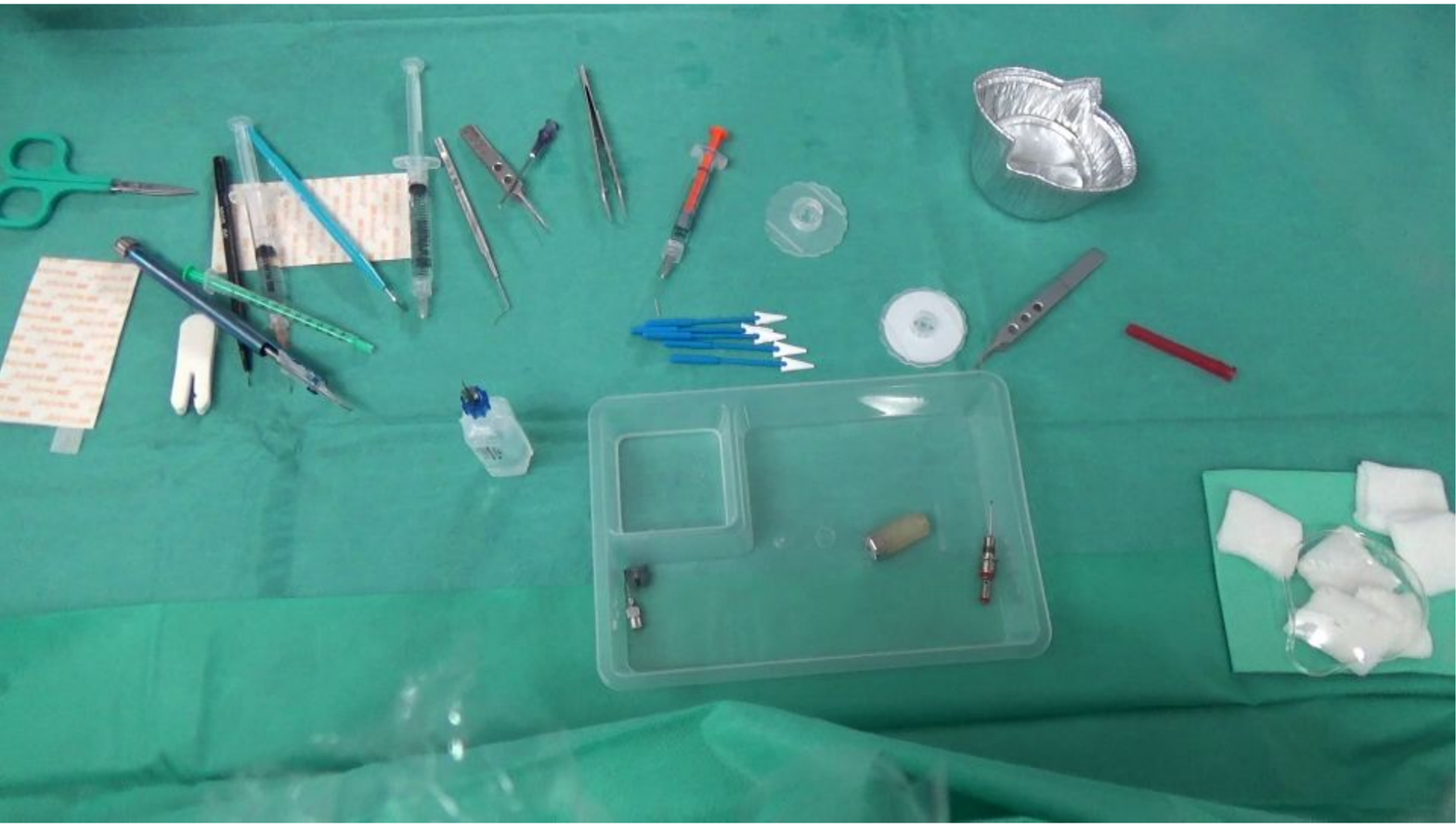}
      } \\
      \subfloat[Manual segmentation of \changed{'before' and 'after' images}]{
        \includegraphics[width=.44\textwidth]{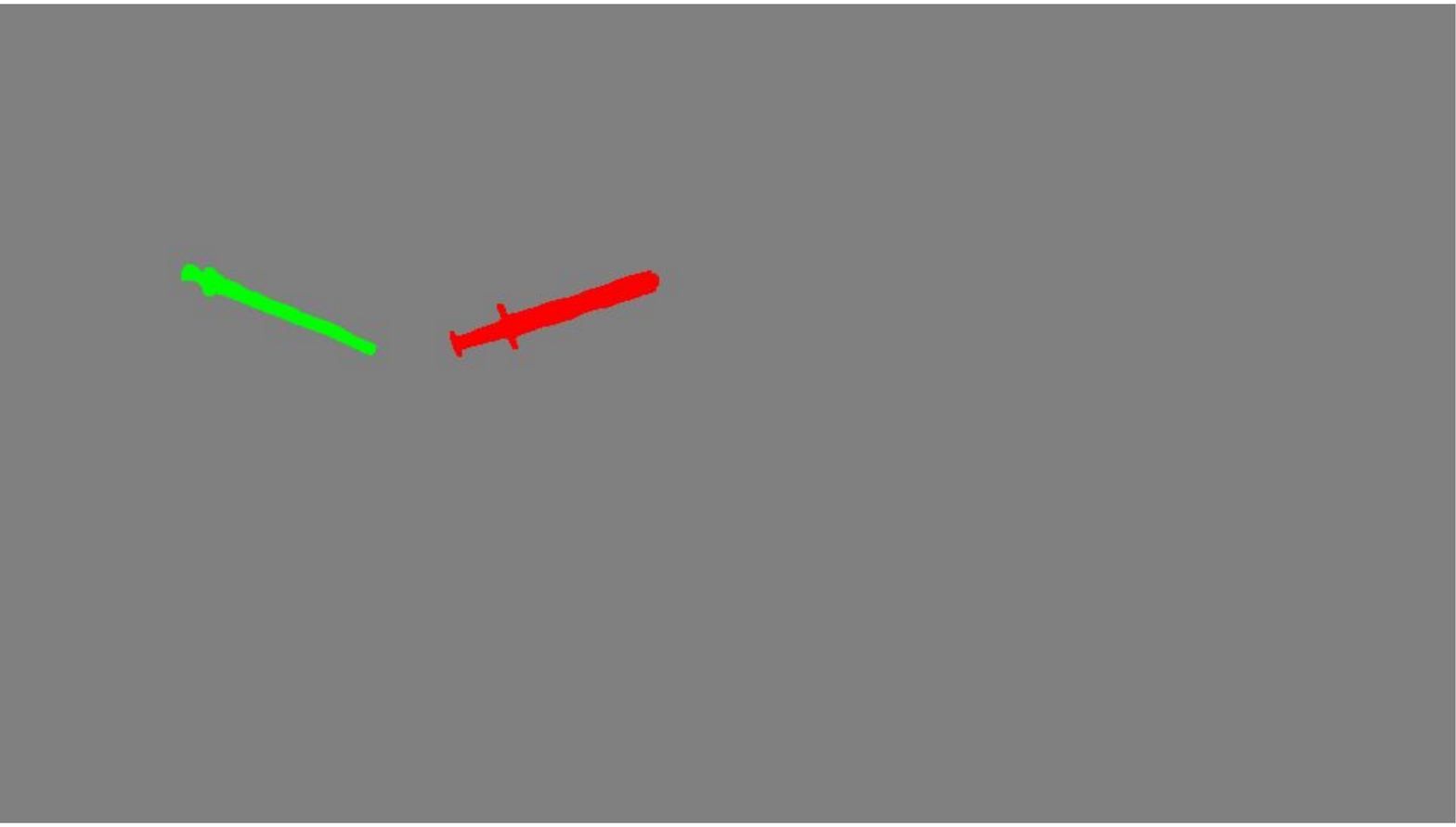}\label{fig:objectDectectionManualSegmentation2}
      } &
      \subfloat[Result of instrument detection]{
        \includegraphics[width=.44\textwidth]{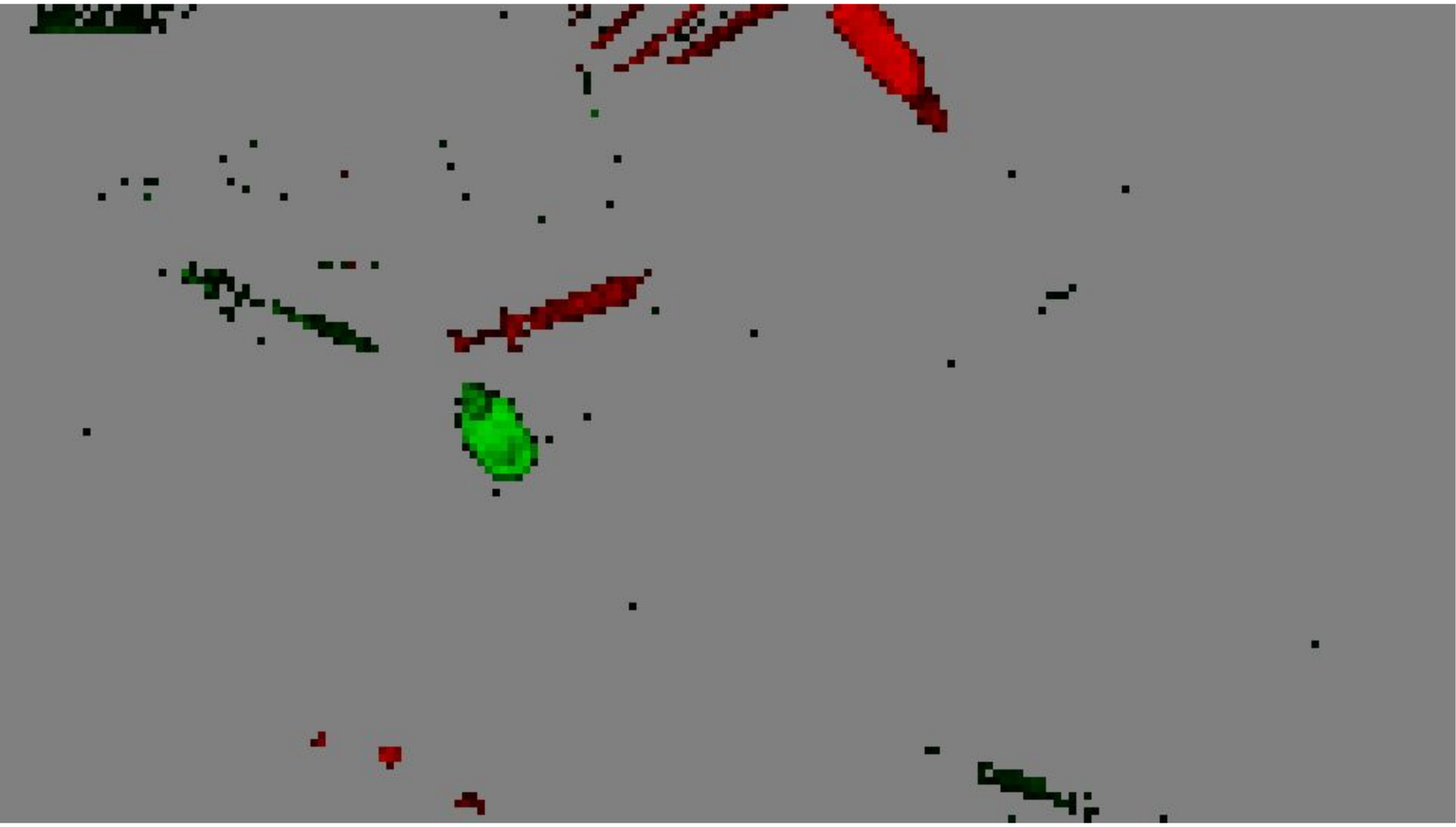} \label{fig:objectDectectionResultsProblems}
      }
    \end{tabular}
  \end{center}
  \caption{Two examples of dynamic instrument detection: \changed{a success and a failure}. In (d), (e), (h) and (i) red indicates the objects that left the table scene and green represents the objects \changed{that entered} the scene. }
  \label{fig:dynamicDetectionResults}
\end{figure}

\begin{table*}
  \caption{Performance $A_{z}$ of static instrument detection}
  \centerline {
    \begin{tabular}{|c|c|c|c|c|c|c|c|}
      \hline
      Type & $K$ & $\tau$ & $P_{min}$ & $P$-levels & $P$-sizes & $A_{z}$ Mean & $A_{z}$ Std \\
      \hline
      \hline
      D-PSO & 89 & 4 & 5 & 3 &  [5;20;80] & 0.982 & 0.015 \\
      \hline
    \end{tabular}
  }
  \label{tab:azStaticInstrumentDetection}
\end{table*}

\begin{table*}
  \caption{Performance $A_{z}$ of dynamic instrument detection}
  \centerline {
    \begin{tabular}{|c|c|c|c|c|c|c|c|c|}
      \hline
      Type & $K$ & $\tau$ & $W$-size & $P_{min}$ & $P$-levels & $P$-sizes & $A_{z}$ Mean & $A_{z}$ Std \\
      \hline
      \hline
       Research on grid & 89 & 4 & 81 & 5 & 3 &  [5;20;80]
 & 0.947 & 0.045  \\
      \hline
    \end{tabular}
  }
  \label{tab:azDynamicInstrumentDetection}
\end{table*} 

\subsubsection{Dynamic Method}
to streamline the optimization process, we assumed that values of the parameters obtained in the previous section can be used in this part. So, we fixed the values of the common parameters and we just optimized $W$-size using a grid search, by randomly assigning values to it. The results are presented in Table \ref{tab:azDynamicInstrumentDetection}. In Fig. \ref{fig:dynamicDetectionResults}, we show that we could identify the instruments that have left and appeared in the scene. But one clear limitation, presented in Fig. \ref{fig:dynamicDetectionResults}\subref{fig:objectDectectionResultsProblems}, is when instruments are seen under a very different view in the 'before' and 'after' images.


\section{Conclusion}
A promising solution to detect the instruments over the operating table has been presented in this paper. \changed{The proposed solution is based on k-NN regression, using a coarse-to-fine strategy. In future works}, more advanced features will be proposed to push performance further. \changed{Also,} to achieve the set aims, we will need to automate the selection of 'before' and 'after' images. The next step will be to recognize the detected objects. A k-NN regression principle can also be followed for this task, possibly using a temporal model of the surgery to help discriminate between strongly resembling instruments. The resulting tool will be very useful for computer-aided surgery.

\bibliographystyle{IEEEtran}
\bibliography{EBMC2016}

\end{document}